\documentclass{article}

\newif\ifanonymous
\anonymousfalse   

\ifanonymous
  \usepackage{neurips_2026}
\else
  \usepackage[preprint]{neurips_2026}
\fi

\usepackage[utf8]{inputenc} 
\usepackage[T1]{fontenc}    
\usepackage{hyperref}       
\usepackage{url}            
\usepackage{booktabs}       
\usepackage{amsfonts}       
\usepackage{nicefrac}       
\usepackage{microtype}      
\usepackage{xcolor}         
\usepackage{svg}
\usepackage{subcaption}
\usepackage{multirow}
\usepackage[table]{xcolor}
\newcommand{\prunemethod}{Putri}

\title{Prune, Update and Trim: Robust Structured Pruning for Large Language Models}

\author{%
  Diego Coello de Portugal Mecke\thanks{\texttt{coellod@uni-hildesheim.de}} \\
  ISMLL \& DARC VWFS\\
  University of Hildesheim\\
  Hildesheim, Germany\\
  \And
  Tom Hanika\\
  ISMLL\\
  University of Hildesheim\\
  Hildesheim, Germany\\
  \And
  Lars Schmidt-Thieme\\
  ISMLL\\
  University of Hildesheim\\
  Hildesheim, Germany\\
}

\begin{document}

\maketitle

\begin{abstract}
Large Language Models (LLMs) have experienced significant growth and development in recent years.
However, performing inference on LLMs remains costly, especially for long-context inference or in resource-constrained devices.
This motivates the development of new post-training pruning (PTP) methods.
These methods reduce LLMs' requirements by removing a substantial part of the model's parameters.
The discarded weights are selected depending on their impact on the models performance.
Current PTP methods prune the models by removing the less informative hidden nodes from the FFN layers,
and the least important attention layers.
We propose \textbf{\prunemethod}, a PTP method that introduces three changes to the State-of-the-art.
First, we update the un-pruned weights of the FFN to compensate for the introduced pruning error.
Second, the FFN layers are pruned sequentially, taking into account the updates done to the previous layers.
Third, instead of removing full attention layers, we remove individual attention-heads.
We extend this method such that it can also address Grouped-Query Attention.
In summary, \prunemethod{} is a structure pruning method which remains simple while showing SOTA performance.
Pruning experiments on multiple models with a wide variety of sparsity ranges and on different datasets, validate the generality of \prunemethod.
Notably, we demonstrate that, unlike previous methods, \prunemethod{} can prune LLMs on extreme sparsity ratios.
\ifanonymous
\else
    The code is available at: https://github.com/Coello-dev/Putri.
\fi
\end{abstract}

\section{Introduction}

Large Language Models (LLMs) \cite{gpt1, gpt2, gpt3} show consistent improvements when increasing model scale, training data and compute.
This enhanced performance has allowed LLMs to tackle a variety of tasks such as agentic workflow \cite{agent1,agent2,agent3} and robotics \cite{rt1,palme,rt2}. Nevertheless, these tasks require the model to either run on a large number of tokens, which drastically increases the required memory, or run on a smaller device with less memory and computing capacity than data centers.
These requirements have motivated the development of multiple post-training techniques that reduce the memory and computing requirements of LLMs by compressing model size.

For example, \textit{Knowledge distillation} \cite{knowledgedistillation} aims to train a smaller model (\textit{student}) that mimics the prediction of the already trained bigger model (\textit{teacher}).
Even though \textit{Knowledge distillation} achieves a highly optimized smaller model, it requires the computational effort to run inference on the \textit{teacher} model and train the \textit{student} model at the same time.
Another approach is \textit{Quantization} \cite{quantization}, which reduces the memory requirements by reducing the floating-point precision of the model's parameters.
However, this technique is limited by the number of precision bits to store the weights.
This sets a lower bound of how much it can compress the model.
Another prominent method is \textit{Pruning} \cite{obd}, which aims to remove the less informative parameters of a model while maintaining the original model performance.
This allows Pruning to reduce the model size with a comparatively small computational effort (with respect to  Knowledge Distillation).
At the same time it is (in theory) capable of reducing the model size as much as required for a certain task (in contrast to Quantization).
Nevertheless, in practice Pruning methods are not used due to its inability to meaningfully reduce the model's size.

The first Pruning methods aimed to remove the model's weights without considering the model's structure (\textit{Unstructured pruning}~\cite{magnitudeprune,sparsegpt}).
These methods result in models with highly sparse weight matrices, i.e., weight matrices where most of the values are zero.
Although these procedures allow for pruning at high sparsity levels without significantly reducing the model's performance, they do not always result in memory or speed gains proportional to the sparsity ratio~\cite{besa,ria}.
This inability to turn the sparsity into meaningful memory and speed-up gains is a result from current software and hardware not being able to benefit from sparse matrices.
In particular, current computation libraries such as (Pytorch \cite{pytorch}, Keras \cite{keras}, etc) and accelerator hardware (GPUs, TPUs, etc.) are highly optimized for dense matrix operation and only partially benefit from sparse matrices.
Even with specialized frameworks \cite{deepsparse}, these Unstructured Pruning methods do not lead to relevant speed-up or memory gains. 
Notably, these marginal gains are not proportional to the model's sparsity \cite{sparsegpt}.

\emph{Structured Pruning} methods~\cite{llmpruner} take a different approach.
Their sparsification methods take into account the underlying model structure.
This guarantees both a reduction of the memory footprint and a substantial speed-up for inference.
A principled way to do this is to remove rows or columns from the weight matrices (\textit{Width pruning}~\cite{slicegpt}).
This reduces the number of computations required while maintaining matrices dense for faster computation.
Another Structured Pruning approach, it to remove full transformer layers (\textit{Depth pruning} \cite{shortgpt}) or any other important substructure such as attention heads~\cite{slimllm}.

State-of-the-art pruning methods such as \emph{2SSP} \cite{2ssp} achieve efficient post-training compression by applying simple criteria.
However, these methods leave several design choices unexplored.
First, they do not update the remaining weights after a pruning step. 
This can lead to a significant reconstruction error at high sparsity.
Second, when pruning attention layers, their approach works with a coarse granularity, i.e., full entire attention modules are removed instead of removing individual attention heads.
Third, pruning multiple FFN layers independently ignores the distribution shift introduced by earlier pruning steps.

We address these limitations by introducing \textbf{\prunemethod}, a robust structured pruning method with the following contributions:
\begin{itemize}
    \item We show the importance of updating the unpruned weights.
    In particular, we demonstrate how some of the state-of-the-art (SOTA) pruning methods underperform for this reason.
    \item Instead of discarding entire attention layers, we introduce
      a fine-grained attention head removal.  Previous attention head
      pruning methods did only consider standard Multi-Head Attention
      (MHA).  To the best of our knowledge, our approach is the first
      method to tackle Grouped-Query Attention (GQA), which is a
      standard component in many modern LLMs.
    \item We sequentially prune the FFN layers, allowing each pruning decision to account for the perturbations introduced by the layers that were pruned earlier.
    \item Our method not only achieves SOTA performance, but it also demonstrates that achieving 95\% sparsity is possible with Structured Pruning.
    This opens the door to explore in the future Structured Pruning methods that could compete with other model compression approaches such as Quantization or Knowledge Distillation. 
\end{itemize}


\section{Related Work}

Pruning is a model compression technique that aims to remove the unimportant weights.
This can be done in an unstructured manner where you remove some of the weights' parameters depending on an importance score.
For example, using the weights' magnitude \cite{magnitudeprune}, or given a calibration dataset,
it can measure some statistics of the weights and use it to score and prune the weights' parameters.
This includes the expected squared error \cite{wanda}, the standard deviation of the output \cite{stade} or minimize the error reconstruction using the Hessian to approximate the loss \cite{sparsegpt}.
Even though Unstructured Pruning methods can maintain good performance even for high sparsity,
they require specialized software to benefit from the sparse matrices \cite{deepsparse}, they don't have reliable speed-ups \cite{slicegpt} and they do not reduce the memory requirements.
This motivates the idea of pruning sub-structures of the model that can reliably lead to memory and speed-up improvements.

\subsection{Structured Pruning}

\textit{Structured Pruning} prunes the model while taking into account the underlying structure.
The different pruning methods can be grouped as follows:

\paragraph{Width pruning} These pruning methods focus on reducing the number of columns and rows of the weights' matrices \cite{slicegpt,2ssp}.
Intuitively, it prunes some of the hidden neurons and/or channels from the model.
\paragraph{Depth pruning} In this case, given a model with skip connections ($x_{l+1} = x_l + f_l(x_l)$) it aims to prune the model by removing the processing blocks $f_l(x_l)$ such that $x_{l+1}=x_l$.
This can be applied to the full transformer layer \cite{shortgpt,uidl} or to the individual sub-blocks, i.e., the FeedForward Layer and the Multi-Head Attention \cite{blockpruner,evopress,2ssp}.
\paragraph{Head Pruning} It prunes the attention layer by removing individual heads, which allows to remove multiple columns from the query, key and value matrices as well as multiple rows from the output matrix.
To the best of our knowledge, existing methods can only be applied to standard Multi-Head Attention (MHA) \cite{slimllm} and not to Grouped-Query Attention (GQA).
Therefore, they can not be used to prune LLMs with GQA such as Llama3 or Qwen3.

\subsection{Grouped-Query Attention}

When performing inference with a transformer model, the process is done in an auto-regressive manner.
This process recomputes many calculations, in particular, the key and values of previous tokens in the attention layer is recomputed for each auto-regressive step.
In order to avoid that, LLMs store the previous tokens' keys and values for faster inference (KV-cache).
However, this only speed-ups the process but does not reduce the memory requirements.
To improve the memory efficiency, Multi-Query Attention (MQA \cite{mqa}) was proposed.
GQA is a variation of MHA that uses only one key and value head that matches all the query-heads.
This also reduces the KV-cache required, making the inference process even faster.

Nevertheless, MQA is not able to perform comparably to standard MHA.
Grouped-Query Attention (GQA \cite{gqa}) is introduced as an option in between MHA and MQA, where each key-value head matches to multiple query heads.
However, contrary to MQA there are multiple key-value heads instead of a single one.
This allows GQA to benefit from the speedup by having less key-value heads than MHA and therefore less KV-cache, while still achieving the same performance as MHA.

\section{Methodology}

Following \textit{2SSP}~\cite{2ssp} and we split LLM pruning process in 2 steps.
First, we prune the MLP by removing some intermediate nodes.
However, we do this process in a sequential manner while additionally update the remaining weights in the process.
Second, we prune the Attention layers,
but instead of removing full attention layers, we remove individual heads from MHA or KV-heads from GQA.

\subsection{MLP pruning}
Given a target sparsity for the MLP layers we prune all the MLP layers uniformly.
Achieving the same sparsity level on all of them.
For each MLP layer, the pruning process aims to remove all the connections related to the less important intermediate nodes/hidden representations.
In order to prune the intermediate nodes, first a score is assigned to each individual node as follows:

\begin{equation}
    score(node_i^{(l)}) = \| z_i^{(l)}\|_2^2
\end{equation}

where $i$ is the node index, $l$ refers to the layer and $z_i^{(l)}$ represents the output of the intermediate node or analogously, the input to the \textit{mlp.fc2} or \textit{mlp.down\_proj}.
After removing the nodes with the least score, each un-pruned connection from the intermediate weights is reconstructed by minimizing the $L_2$-loss, i.e.:

\begin{equation}
    arg \min_{\hat{W}_P} \|XW - X_P \hat{W}_P\|_2^2 = (X_P^T X_P)^{-1} X_P^T XW
\end{equation}

where $X \in \mathbb{R}^{N \times M_1}$ and $W \in \mathbb{R}^{M_1 \times M_2}$ are the original inputs and weight parameters, while $X_P \in \mathbb{R}^{N \times P}$ and $\hat{W}_P \in \mathbb{R}^{P \times M_2}$ are the inputs of only the $P$ unpruned intermediate nodes and their corresponding updated weights.
$N$ refers to the number of instances in the dataset, $M_1$ the number of input features and $M_2$ the number of output features.
Applying this update, allows the remaining weights to reduce some of the error obtained from pruning some intermediate representations.

In order to further improve performance, the MLP layers are not pruned in parallel but sequentially.
This allows each MLP layer to be pruned taking into account the output of the previously pruned layers.

\begin{figure}[t]
    \centering
    \begin{subfigure}[t]{0.46\linewidth}
        \centering
        \includegraphics[width=\linewidth]{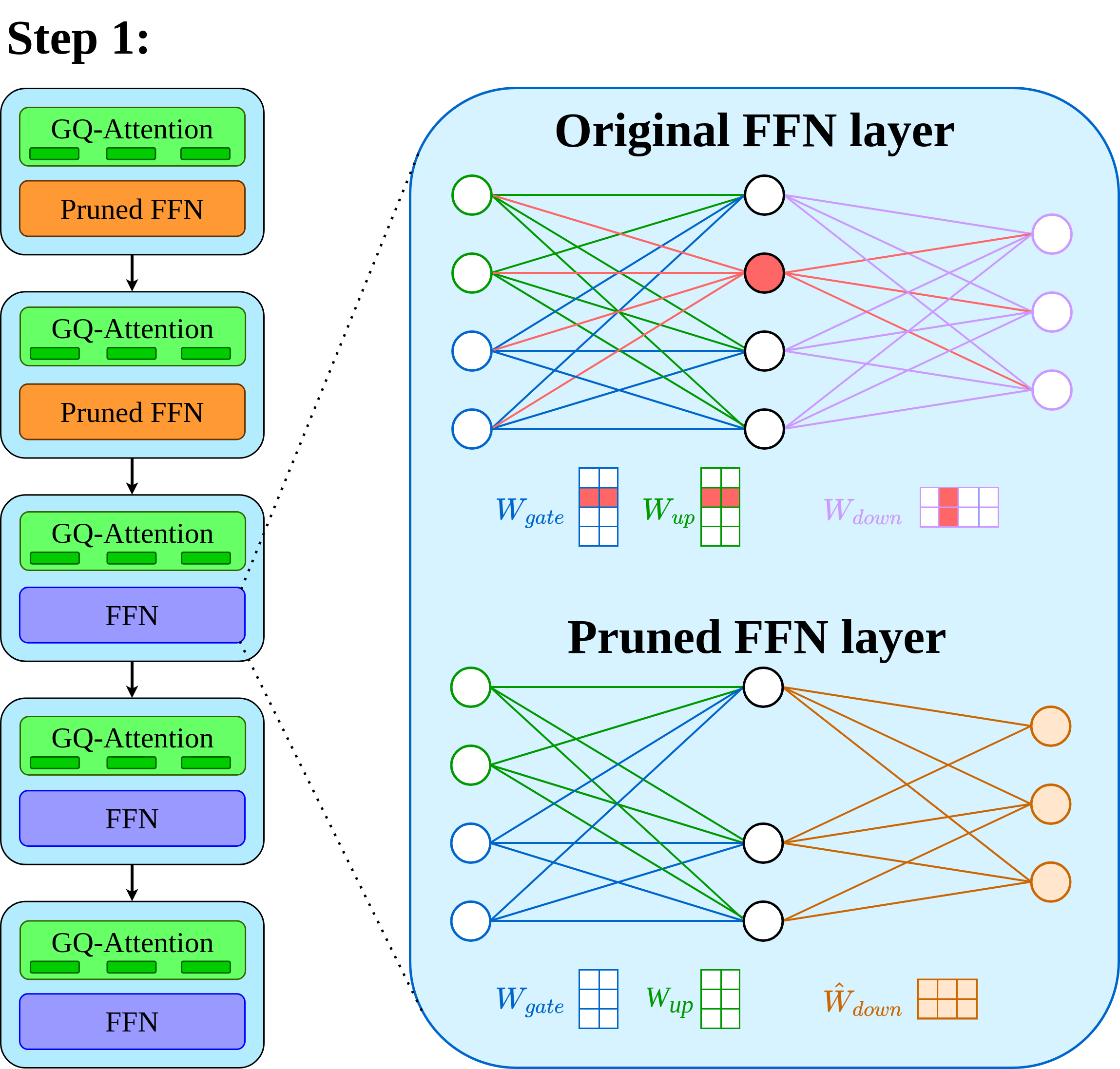}
    \end{subfigure}\hfill
    \begin{subfigure}[t]{0.51\linewidth}
        \centering
        \includegraphics[width=\linewidth]{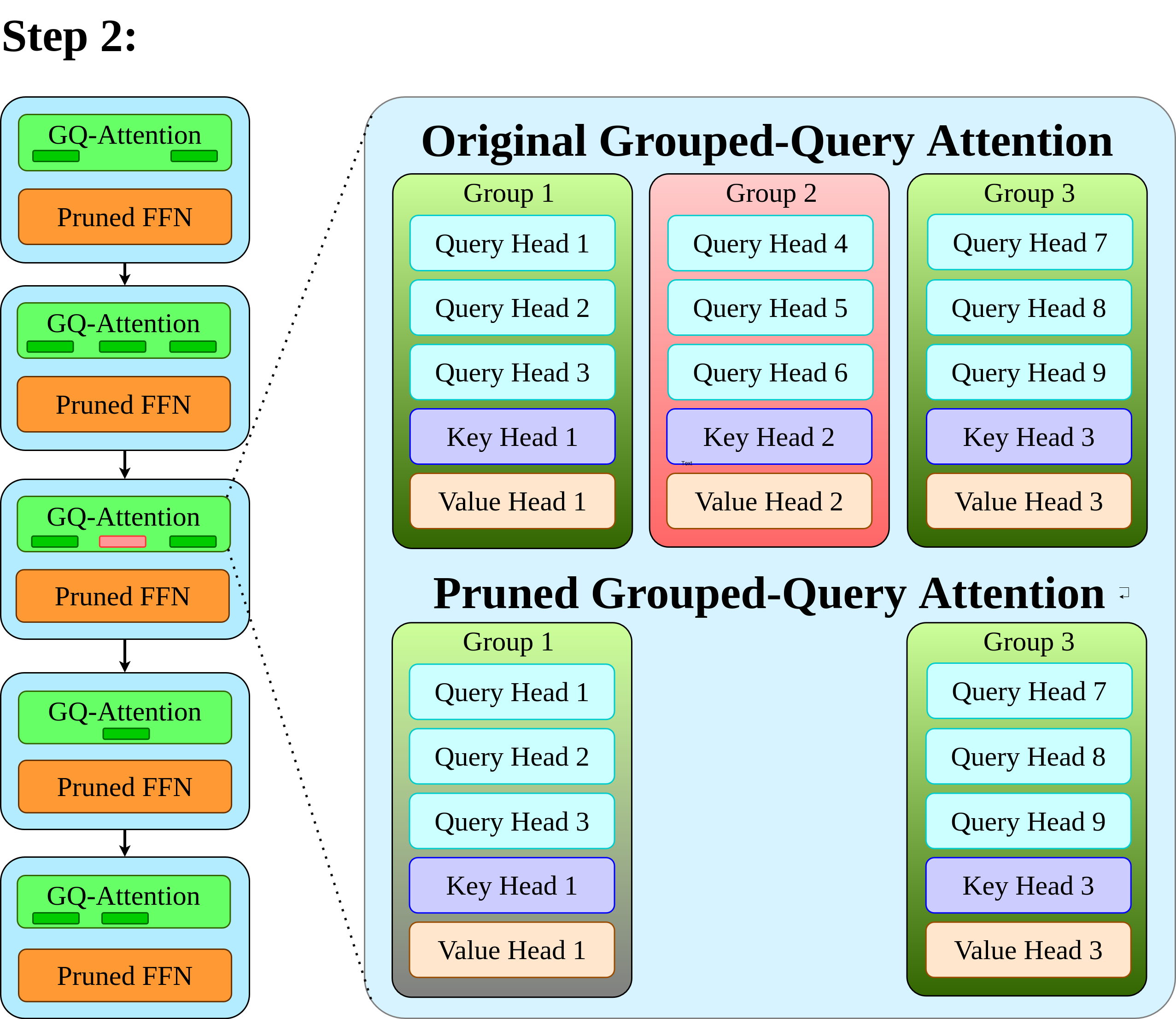}
    \end{subfigure}
    \caption{Diagram of the pruning process for SPU-H.
    First the FFN layers are pruned sequentially.
    Then the attention/kv-heads are pruned in an iterative manner.}
\end{figure}

\subsection{Attention pruning}

Previous works~\cite{evopress} showed that deciding which full attention layers to remove using the perplexity on 1 sequence achieves comparable performance to using multiple sequences.
This reduces dramatically the time required to prune.
However, we argue that a more fine-grained approach, such as pruning each head, could lead to a better performance.
Notice that some of the modern LLMs pruned use Grouped-Head Attention instead of Multi-Head Attention (MHA).
We address this challenge by selecting the individual key-value heads to prune and all their associated q-heads.
In the case of MHA, each key-value head has one unique query head associated, while for GQA there would be multiple query heads for each key-value head.
In particular we select the key-value head to remove as follows:

\begin{equation}
    arg \min_{a \in \theta_{Attn}} perplexity(f(X; \theta_{FFN}, \theta_{Attn}\backslash\{a\}), Y)
\end{equation}

where  $X \in \mathbb{R}^{N \times M_1}$ are the inputs, $\theta_{FFN}$ are the weight parameters from the FFN layers, $\theta_{Attn}$ are the weight parameters from the Attention layers
and $a$ refers to a grouped head from $\theta_{Attn}$, including multiple query heads, a key head and a value head.
We notice that selecting one grouped head at a time and recalculating the perplexity each time requires a lot of compute.
We address this limitation by removing the H less informative heads each iteration, where H is the number of grouped heads in each of the original attention layer.

\subsection{Attention vs FFN Pruning ratio}

Previous methods \cite{2ssp} showed that the ideal pruning ratio between the Attention and FFN layers is obtained by pruning a number of attention layers equal to:

\begin{equation}
    \label{eq:ratio_ffn_attn}
    N_{Attn} = round(L \cdot s^{\frac{|\theta_{FFN}|}{\alpha |\theta_{Attn}|}})
\end{equation}

where $L$ is the number of layers, $s$ is the target sparsity, $|\theta_{i}|$ is the number of parameters that the $i$ part of the model has (with $i \in \{FFN, Attn\}$) and $\alpha$ is the hyperparameter controlling the ratio ($\alpha=1.5$ by default).
In our case, since we are pruning key heads instead of attention layers, we modify Equation \ref{eq:ratio_ffn_attn} slightly to calculate the number of key-value heads as follows:

\begin{equation}
    \label{eq:ratio_ffn_attn_heads}
    N_{KV-Heads} = K \cdot round(L \cdot s^{\frac{|\theta_{FFN}|}{\alpha |\theta_{Attn}|}})
\end{equation}

where $K$ is the number of key heads in each Grouped-Query Attention layer.

\section{Experiments}

Previous Structured Pruning methods evaluate at a moderate sparsity range with maximum ratio of 50\% sparsity.
However, the final objective of pruning is to \textbf{significantly} reduce the number of parameters from the model, i.e.,
reduce the model size by at least an order of magnitude.
This reduction would require of a sparsity ratio of at least 90\%.
Nevertheless, this scenario remains unexplored in the literature with no discussion or experimentation.
In this section we will aim to not only show the performance of our method (\textbf{\prunemethod}) but also fill in the gap in the literature regarding this sparsity range.

\subsection{Set-Up}

In order to do a fair comparison between all the pruning method, we use an extensive set-up with a wide variety of model families, model sizes, datasets with different tasks, pruning methods and sparsities ranges.
The specific configurations used in the experiments consider the following options:

\paragraph{Models} We evaluate the Llama-3 \cite{llama3} (including 3.0, 3.1 and 3.2 versions) and Qwen3 \cite{qwen3} family of models across the different sizes available.
Both of them are SOTA open-sourced modern LLMs with GQA \cite{gqa}, RMSNorm \cite{rmsnorm} and RoPE embeddings \cite{roformer}.

\paragraph{Datasets} For calibration during the pruning process we use the \textit{C4} dataset \cite{c4dataset}.
The pruned models' perplexity is evaluated on the datasets \textit{C4} \cite{c4dataset}, \textit{Wikitext-2} (Wiki \cite{wikitext2}) and \textit{FineWeb-Edu} (FW \cite{fineweb}).
Additionally, we also measure the zero-shot capabilities using LM Eval Harness \cite{eval-harness}.
We evaluate on multiple benchmarks such as \textit{Winogrande} \cite{ai2:winogrande}, \textit{ARC} (easy and challenge, "ARC-E" and "ARC-C" respectively) \cite{allenai:arc}, \textit{Hellaswag} \cite{zellers2019hellaswag}, \textit{PiQA} \cite{piqa} and \textit{MMLU} \cite{mmlu}.

\paragraph{Baselines} We evaluate against multiple post-training Strcutured Pruning methods,
which include width-wise pruning approaches (\textit{2SSP}~\cite{2ssp}) and depth-wise pruning on the transformer blocks (\textit{ShortGPT}~\cite{shortgpt}, \textit{UIDL}~\cite{uidl}) or the FFN and Attention sub-blocks (\textit{Blockpruner}~\cite{blockpruner},  \textit{Evopress}~\cite{evopress}, \textit{2SSP}~\cite{2ssp}).
Since most new open-sourced LLMs do not use MHA but GQA, we do not evaluate previous head pruning methods since they are not suitable for GQA.

\paragraph{Sparsity} As previously mentioned, we will focus on higher sparsity ranges.
Therefore we would start from 50\% sparsity and prune until a sparsity of 99\%.

Due to space constrains, we delegate to the Appendix the following results.
First, the perplexity results on the test partition of the C4 dataset.
We consider Wikitext-2 and FineWeb-Edu more interesting datasets since the models are pruned with the train partition of C4.
Therefore, this other datasets would better show the generalization to text not following the distribution of the calibration dataset C4.
Second, the smaller models from the Llama3 family. Qwen3 has shown in general better capabilities than Llama3 and therefore we only show results on the bigger Llama3 models we are able to run.
Lastly, we measure for completeness the performance of the pruning methods in lower sparsities such as 25\% and 37.5\%.

\begin{table}[t]
\caption{Perplexity comparison of multiple pruning methods on Qwen3 and LLama3 models.
We add some notations to specify the following cases: `OOM' is an out-of-memory error, `nan' refers to an invalid pruned model output or an error during the pruning process, `$\infty$' is a successfully pruned model that returns infinite when the perplexity is evaluated.}
\label{tab:perplexity_main}
\centering
\begin{tabular}{llcccccc}
\toprule
\multirow{2.5}*{Sparsity} & \multirow{2.5}*{Method} & \multicolumn{2}{c}{Llama-3.1-8B} & \multicolumn{2}{c}{Qwen3-14B} & \multicolumn{2}{c}{Qwen3-32B} \\
\cmidrule(lr){3-4} \cmidrule(lr){5-6} \cmidrule(lr){7-8}
 &  & Wiki & FW & Wiki & FW & Wiki & FW \\
\midrule
0.0 & - & 6.406 & 8.375 & 8.641 & 9.43 & 7.594 & 8.32 \\
\midrule
\multirow{6}*{0.5} & UIDL & 52736 & 38656 & $\infty$ & $\infty$ & $\infty$ & $\infty$ \\
 & ShortGPT & 14208 & 10432 & 1318 & 622.5 & $\infty$ & $\infty$ \\
 & BlockPruner & 163 & 163 & 221 & 131.5 & 39.25 & 35.03 \\
 & EvoPress & 148 & 96 & 86.94 & 73.75 & OOM & OOM \\
 & 2SSP & \textbf{34.75} & \textbf{41.25} & \underline{23.53} & \underline{25} & \underline{19.39} & \underline{21.67} \\
\rowcolor{gray!50} \cellcolor{white}  & Putri (ours) & \underline{39.25} & \underline{41.75} & \textbf{18.19} & \textbf{19.97} & \textbf{15.16} & \textbf{16.95} \\
\midrule
\multirow{6}*{0.75}& UIDL & 344064 & 196608 & $\infty$ & $\infty$ & $\infty$ & $\infty$ \\
 & ShortGPT & 49664 & 49664 & $\infty$ & $\infty$ & 64736 & $\infty$ \\
 & BlockPruner & 98816 & 28288 & 5524 & 4336 & 5884 & 3224 \\
 & EvoPress & 17152 & 8640 & 7208 & 4840 & OOM & OOM \\
 & 2SSP & \underline{278} & \underline{260} & \underline{212.6} & \underline{160.5} & \underline{122.1} & \underline{99.62} \\
\rowcolor{gray!50} \cellcolor{white}  & Putri (ours) & \textbf{163} & \textbf{185} & \textbf{124.5} & \textbf{68.75} & \textbf{44.66} & \textbf{41.78} \\
\midrule
\multirow{6}*{0.9}& UIDL & 1859584 & 21364736 & $\infty$ & $\infty$ & nan & nan \\
 & ShortGPT & 1859584 & 21364736 & $\infty$ & $\infty$ & nan & nan \\
 & BlockPruner & 501760 & 684032 & nan & nan & 26992 & 27840 \\
 & EvoPress & 28288 & 14208 & 9928 & 7208 & OOM & OOM \\
 & 2SSP & \underline{1544} & \underline{908} & \underline{5316} & \underline{2662} & \underline{2866} & \underline{1476} \\
\rowcolor{gray!50} \cellcolor{white}  & Putri (ours) & \textbf{644} & \textbf{416} & \textbf{1307} & \textbf{716.5} & \textbf{370.2} & \textbf{215.1} \\
\midrule
\multirow{6}*{0.95}  & UIDL & 501760 & 501760 & $\infty$ & $\infty$ & \underline{$\infty$} & \underline{$\infty$} \\
 & ShortGPT & 532480 & 391168 & $\infty$ & $\infty$ & $\infty$ & $\infty$ \\
 & BlockPruner & 1204224 & 995328 & nan & nan & nan & nan \\
 & EvoPress & 135168 & 98816 & \underline{10816} & \underline{8360} & OOM & OOM \\
 & 2SSP & \underline{8096} & \underline{2976} & nan & nan & \underline{29872} & \underline{23264} \\
\rowcolor{gray!50} \cellcolor{white}  & Putri (ours) & \textbf{1696} & \textbf{1096} & \textbf{4916} & \textbf{1880} & \textbf{2714} & \textbf{871} \\
\midrule
\multirow{6}*{0.99} & UIDL & nan & nan & nan & nan & nan & nan \\
 & ShortGPT & 532480 & 501760 & \underline{$\infty$} & \underline{$\infty$} & \textbf{$\infty$} & \textbf{$\infty$} \\
 & BlockPruner & 532480 & 501760 & nan & nan & nan & nan \\
 & EvoPress & nan & nan & nan & nan & OOM & OOM \\
 & 2SSP & \underline{34048} & \underline{17152} & nan & nan & nan & nan \\
\rowcolor{gray!50} \cellcolor{white} & Putri (ours) & \textbf{11776} & \textbf{5216} & \textbf{46624} & \textbf{39584} & $\infty$ & $\infty$ \\
\bottomrule
\end{tabular}
\end{table}

\subsection{Main experiment}

Even though there are speed-up and memory benefits when pruning a model up to 50\%,
those benefits need to be of at least an order of magnitude in order to be deployed in real world scenarios.
A speed-up or memory reduction of an order of magnitude can only be achieved with a sparsity 90\% or higher.
Given this reason, we focus our evaluation of the pruning methods on sparsity levels from 50\% sparsity up to 99\%.
To accurately measure the generalization capabilities of the pruning methods, we evaluate the perplexity over many datasets and multiple models in Table \ref{tab:perplexity_main}.

The consistent improvements in Table \ref{tab:perplexity_main} show the robustness of our method compared to the baselines.
Additionally, we also show that most pruning methods are not able to properly prune on high sparsity degrees.
In particular, they either return models with infinite perplexity (denoted with $\infty$) or the pruning method is not able to successfully finish due to not being able to handle such a high sparsity (denoted with `$nan$').

To further investigate the collapse of the pruning methods, we evaluate them on a wider range of model sizes from the Qwen3 family of models.
We particularly focus on the smaller models since achieving a higher sparsity degree on them is an even harder task compared to the bigger models.
The results in Table \ref{tab:perplexity_growth} show that not only \prunemethod{} improvements are consistent across the model sizes,
but also it is able to successfully prune to a much higher sparsity degree than previous methods.
Nevertheless, pruning the smaller models with a pruning sparsity of 99\% is not even possible for \prunemethod.
This further highlights the difficulty of pruning in this highly sparse ranges.
The complete list of results for all the different models and set-ups can be viewed in the Appendix.

\begin{table}[t]
\caption{Perplexity comparison across different model sizes.
We add some notations to specify the following cases: `OOM' is an out-of-memory error, `nan' refers to an invalid pruned model output or an error during the pruning process, `$\infty$' is a successfully pruned model that returns infinite when the perplexity is evaluated.}
\label{tab:perplexity_growth}
\centering
\begin{tabular}{llcccccccc}
\toprule
\multirow{2.5}*{Spars.} & \multirow{2.5}*{Method} & \multicolumn{2}{c}{Qwen3-0.6B} & \multicolumn{2}{c}{Qwen3-1.7B} & \multicolumn{2}{c}{Qwen3-4B} & \multicolumn{2}{c}{Qwen3-8B} \\
\cmidrule(lr){3-4} \cmidrule(lr){5-6} \cmidrule(lr){7-8} \cmidrule(lr){9-10}
 &  & Wiki & FW & Wiki & FW & Wiki & FW & Wiki & FW \\
\midrule
0 & - & 20.97 & 20.69 & 16.75 & 16.11 & 13.75 & 13.54 & 9.734 & 10.48 \\
\midrule
\multirow{6}*{0.5} & UIDL & $\infty$ & $\infty$ & $\infty$ & $\infty$ & 3768 & 2946 & $\infty$ & $\infty$ \\
 & ShortGPT & $\infty$ & $\infty$ & 58944 & 11072 & 2570 & 1499 & 1787 & 639.5 \\
 & BlockPruner & 409.8 & 328 & 923.5 & 446.5 & 143.2 & 120.7 & 202.9 & 108.2 \\
 & EvoPress & \underline{356} & 272 & 7792 & 5976 & 117.9 & 100 & 211 & 120.7 \\
 & 2SSP & 371.8 & \underline{255.4} & \underline{166.2} & \underline{109.9} & \underline{63.84} & \underline{56.56} & \underline{41.94} & \underline{40.09} \\
\rowcolor{gray!50} \cellcolor{white}  & Putri (ours) & \textbf{108.2} & \textbf{97.69} & \textbf{65.62} & \textbf{55.47} & \textbf{40.97} & \textbf{36.44} & \textbf{28.44} & \textbf{27.3} \\
\midrule
\multirow{6}*{0.75} & UIDL & $\infty$ & $\infty$ & $\infty$ & $\infty$ & $\infty$ & $\infty$ & $\infty$ & $\infty$ \\
 & ShortGPT & $\infty$ & $\infty$ & $\infty$ & $\infty$ & $\infty$ & $\infty$ & $\infty$ & $\infty$ \\
 & BlockPruner & 20528 & 13256 & $\infty$ & $\infty$ & 14560 & 10088 & 13464 & 5884 \\
 & EvoPress & 6716 & 4952 & 9544 & 6984 & 23264 & 14904 & 5112 & 3768 \\
 & 2SSP & \underline{2866} & \underline{2198} & \underline{1080} & \underline{768.5} & \underline{1596} & \underline{1063} & \underline{536.5} & \underline{395.8} \\
\rowcolor{gray!50} \cellcolor{white}  & Putri (ours) & \textbf{1441} & \textbf{1038} & \textbf{884.5} & \textbf{647} & \textbf{309.2} & \textbf{207.6} & \textbf{164.2} & \textbf{140} \\
\midrule
\multirow{6}*{0.9} & UIDL & \underline{$\infty$} & $\infty$ & $\infty$ & $\infty$ & $\infty$ & $\infty$ & $\infty$ & $\infty$ \\
 & ShortGPT & \underline{$\infty$} & $\infty$ & $\infty$ & $\infty$ & $\infty$ & $\infty$ & $\infty$ & $\infty$ \\
 & BlockPruner & nan & nan & nan & nan & nan & nan & nan & nan \\
 & EvoPress & \underline{$\infty$} & \underline{64736} & 51616 & \underline{23824} & \underline{58496} & \underline{54528} & 34112 & 27408 \\
 & 2SSP & nan & nan & \underline{30816} & 57568 & nan & nan & \underline{12352} & \underline{9256} \\
\rowcolor{gray!50} \cellcolor{white}  & Putri (ours) & \textbf{4440} & \textbf{2414} & \textbf{2610} & \textbf{1685} & \textbf{916.5} & \textbf{705.5} & \textbf{1030} & \textbf{617.5} \\
\midrule
\multirow{6}*{0.95} & UIDL & nan & nan & nan & nan & nan & nan & nan & nan \\
 & ShortGPT & \textbf{$\infty$} & \textbf{$\infty$} & \underline{$\infty$} & \underline{$\infty$} & \underline{$\infty$} & \underline{$\infty$} & \underline{$\infty$} & \underline{$\infty$} \\
 & BlockPruner & nan & nan & nan & nan & nan & nan & nan & nan \\
 & EvoPress & \textbf{$\infty$} & \textbf{$\infty$} & \underline{$\infty$} & \underline{$\infty$} & \underline{$\infty$} & \underline{$\infty$} & \underline{$\infty$} & \underline{$\infty$} \\
 & 2SSP & nan & nan & nan & nan & nan & nan & \underline{$\infty$} & \underline{$\infty$} \\
\rowcolor{gray!50} \cellcolor{white}  & Putri (ours) & \textbf{$\infty$} & \textbf{$\infty$} & \textbf{9400} & \textbf{6068} & \textbf{3596} & \textbf{2258} & \textbf{3124} & \textbf{1540} \\
\bottomrule
\end{tabular}
\end{table}

\subsection{Zero-shot}

Even though perplexity is a good measure on the model's performance, the main goal is for the LLM to be able to correctly answer and solve different tasks.
To evaluate that, we compare the model on a list of benchmarks: Winogrande (WG), ARC-E, ARC-C, HellaSwag (HS), PiQA and MMLU.
The results on Qwen3-14B are shown in Table \ref{tab:zero_shot_qwen3_14b}.

In Table \ref{tab:zero_shot_qwen3_14b}, it can be observed that \prunemethod{} is overall the best method.
However, in those cases were it is the second best (90\% and 99\%) it can be observed in Table \ref{tab:perplexity_main}, that the best method (UIDL) has actually infinite perplexity.
Meaning, that the ``good'' performance in the zero-shot benchmark is a lucky guess rather than an informed decision.
These can also be observed by the fact that at high perplexities all the methods converge to a result close to random.
Notice that most of these benchmarks and questions with a set of options to choose from.
Therefore, even a relatively high accuracy of 33\% or 25\% is nothing more than a random guess out of 3 or 4 options.
This observation highlights again the big gap in the literature when pruning at high sparsities.

\begin{table}
\centering
\caption{Zero-shot accuracy comparison with Qwen3-14B.
`Avg. Acc.' refers to the average accuracy over all the zero-shot tasks and
`nan' refers to an invalid pruned model output or an error during the pruning process}
\label{tab:zero_shot_qwen3_14b}
\begin{tabular}{llccccccc}
\toprule
Sparsity & Method & WG & ARC-E & ARC-C & HS & PiQA & MMLU & Avg. Acc. \\
\midrule
0 & - & 72.8 & 84.2 & 58.7 & 61.0 & 80.0 & 77.3 & 72.3 \\
\midrule
\multirow{6}*{0.5} & UIDL & 52.8 & 28.3 & \textbf{24.4} & 28.0 & 56.7 & 23.0 & 35.5 \\
 & ShortGPT & 50.7 & 30.5 & 19.3 & 26.4 & 53.6 & 23.5 & 34.0 \\
 & BlockPruner & 53.0 & 38.3 & 22.0 & 30.2 & 56.4 & \textbf{25.5} & 37.6 \\
 & EvoPress & 48.5 & 41.1 & 22.2 & 31.8 & 61.3 & \underline{23.9} & 38.1 \\
 & 2SSP & \underline{58.4} & \underline{46.6} & 22.7 & \underline{39.6} & \underline{66.9} & 23.6 & \underline{43.0} \\
\rowcolor{gray!50} \cellcolor{white}   & Putri (ours) & \textbf{62.0} & \textbf{48.8} & \underline{24.0} & \textbf{42.3} & \textbf{69.7} & 23.8 & \textbf{45.1} \\
\midrule
\multirow{6}*{0.75} & UIDL & 48.8 & 26.3 & 22.1 & 25.8 & 54.0 & \textbf{25.9} & 33.8 \\
 & ShortGPT & 47.8 & 24.8 & \underline{22.4} & 25.7 & 51.7 & \underline{25.3} & 33.0 \\
 & BlockPruner & 49.0 & 26.3 & \textbf{23.4} & 25.9 & 53.0 & 24.7 & 33.7 \\
 & EvoPress & \textbf{51.6} & 24.6 & 22.0 & 25.6 & 53.4 & 23.9 & 33.5 \\
 & 2SSP & \underline{49.9} & \underline{28.9} & 17.1 & \underline{28.3} & \underline{56.3} & 24.2 & \underline{34.1} \\
\rowcolor{gray!50} \cellcolor{white}   & Putri (ours) & 49.0 & \textbf{33.1} & 17.7 & \textbf{28.6} & \textbf{56.3} & 23.0 & \textbf{34.6} \\
\midrule
\multirow{6}*{0.9} & UIDL & \textbf{53.4} & 25.6 & \textbf{23.2} & 25.5 & \textbf{53.8} & \underline{24.3} & \textbf{34.3} \\
 & ShortGPT & 50.3 & 24.7 & \underline{22.3} & 25.7 & 52.6 & 22.9 & 33.1 \\
 & BlockPruner & nan & nan & nan & nan & nan & nan & nan \\
 & EvoPress & 48.3 & 24.6 & 21.3 & 25.8 & 52.2 & 23.2 & 32.6 \\
 & 2SSP & 50.3 & \underline{26.3} & 21.4 & \underline{26.1} & 52.6 & 23.4 & 33.3 \\
\rowcolor{gray!50} \cellcolor{white}   & Putri (ours) & \underline{51.6} & \textbf{26.5} & 20.6 & \textbf{26.2} & \underline{53.0} & \textbf{25.4} & \underline{33.9} \\
\midrule
\multirow{6}*{0.95} & UIDL & \textbf{51.1} & \textbf{25.8} & 21.5 & \underline{25.5} & 52.3 & \underline{25.1} & 33.6 \\
 & ShortGPT & 49.4 & 24.5 & \underline{22.8} & 25.2 & \underline{52.6} & \textbf{25.2} & 33.3 \\
 & BlockPruner & nan & nan & nan & nan & nan & nan & nan \\
 & EvoPress & \textbf{51.1} & \underline{25.6} & \underline{22.8} & 25.4 & 51.8 & 24.6 & \underline{33.6} \\
 & 2SSP & nan & nan & nan & nan & nan & nan & nan \\
\rowcolor{gray!50} \cellcolor{white}   & Putri (ours) & \underline{49.5} & 25.3 & \textbf{23.2} & \textbf{26.1} & \textbf{53.5} & 24.5 & \textbf{33.7} \\
\midrule
\multirow{6}*{0.99} & UIDL & nan & nan & nan & nan & nan & nan & nan \\
 & ShortGPT & \textbf{50.8} & \underline{24.3} & \textbf{22.8} & \underline{25.4} & \underline{53.0} & \textbf{26.9} & \textbf{33.9} \\
 & BlockPruner & nan & nan & nan & nan & nan & nan & nan \\
 & EvoPress & nan & nan & nan & nan & nan & nan & nan \\
 & 2SSP & nan & nan & nan & nan & nan & nan & nan \\
\rowcolor{gray!50} \cellcolor{white}   & Putri (ours) & \textbf{50.8} & \textbf{25.7} & \underline{20.8} & \textbf{25.7} & \textbf{53.3} & \underline{24.4} & \underline{33.4} \\
\bottomrule
\end{tabular}
\end{table}

\subsection{Ablation study}

\begin{figure}[h]
    \centering
    \includegraphics[width=\linewidth]{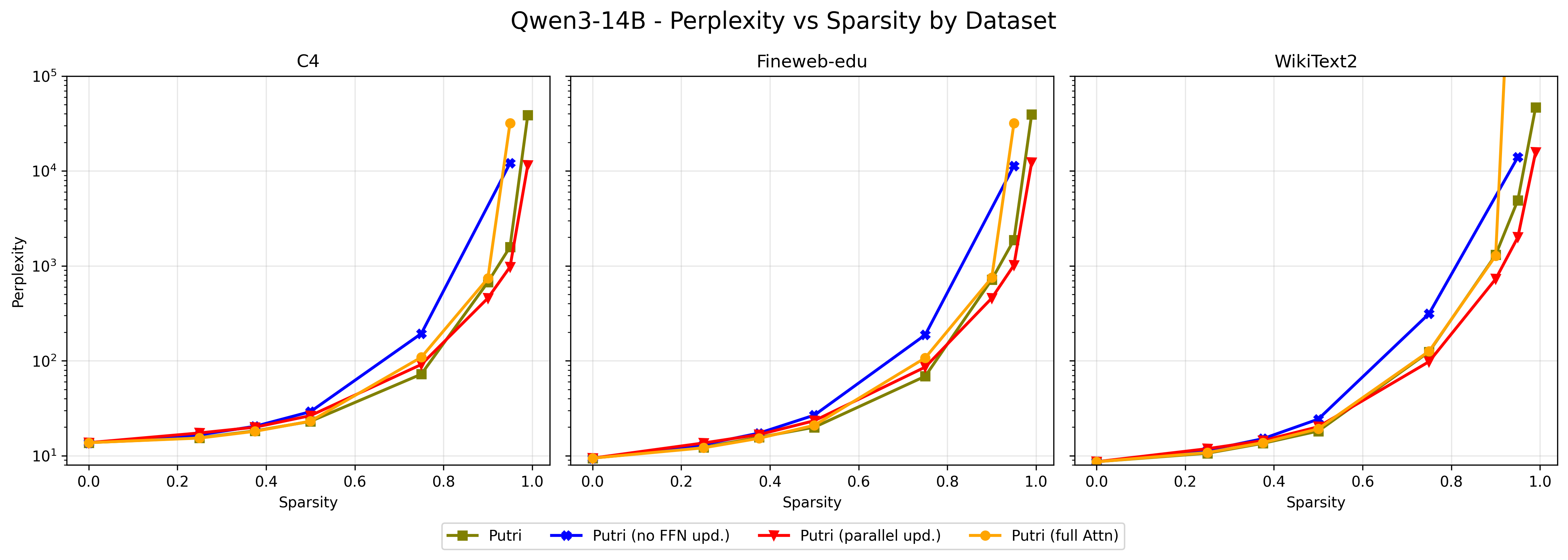}
    \caption{Ablation study on Qwen3-14B about the individual components of \prunemethod.}
    \label{fig:ablation_qwen3_14b}
\end{figure}

\begin{figure}[h]
    \centering
    \includegraphics[width=\linewidth]{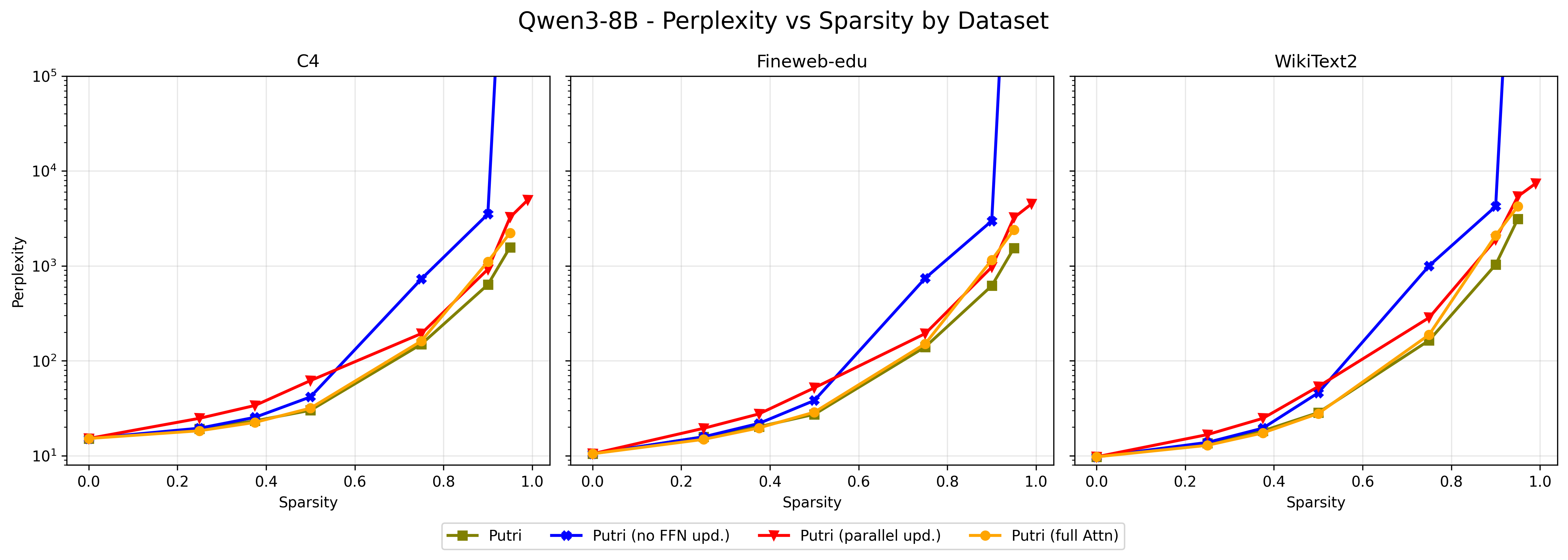}
    \caption{Ablation study on Qwen3-8B about the individual components of \prunemethod.}
    \label{fig:ablation_qwen3_8b}
\end{figure}

Our method \prunemethod{} incorporates 3 main components when compared to previous methods such as 2SSP:
updating the weights of the FFN layers, pruning the FFN layers sequentially to take into account the previous layers pruning, and pruning attention heads instead of the full attention layer.
To validate the usefulness of these additions, we perform the ablation experiments in Figures \ref{fig:ablation_qwen3_14b} and \ref{fig:ablation_qwen3_8b} where each individual component is removed.
We denote `Putri (no FFN upd.)' to not updating the un-pruned FFN weights after removing the other weights,
`Putri (full Attn.)' to remove the full attention layer instead of the heads,
and `Putri (parallel upd.)' to not taking into account the result of the previously pruned FFN layers by using the output of the original model as an input, instead of the already pruned previous layers.
Whenever the pruning method was not able to properly prune the model and resulted in a `nan' error, it is not plotted in the graph.
This happens for some methods at 95\% and/or 99\% sparsity.

We observe in Figures \ref{fig:ablation_qwen3_14b} and \ref{fig:ablation_qwen3_8b} that overall, removing any of these changes results in a decrease in performance.
In particular, not updating the FFN weights significantly reduces the performance, especially for higher sparsities.
We also observe that at even though some of the other combinations can in particular scenarios compete and sometimes even beat the full method,
the overall trend is that the complete method performs better than removing any of the individual parts.

\section{Conclusion}

In this paper we introduce \prunemethod, a post-training structure pruning approach. Our method can even be applied in high-sparsity regimes.
In contrast to previous SOTA methods, \prunemethod{}introduces three main changes.
First, it updates the un-pruned parameters in the last linear layer from the FFN to reduce  errors induced by earlier steps.
Second, the FFN pruning is done sequentially so that each layer takes into account the input from the previously pruned FFN layers.
This not only improves performance but also reduces the memory requirements since only one layer at a time needs to be in memory.
Third, our method takes into account the fact some attention layers depend highly on specific heads rather than in the combination of all of them.
Therefore, \prunemethod{} prunes on a attention head level and extends this approach to tackle Grouped-Query Attention.

Multiple experiments on the Llama3 and Qwen3 family models validate our approach.
We compare against multiple pruning methods in a wide range of sparsities and multiple datasets using perplexity performance.
Additionally, the zero-shot performance is evaluated on multiple benchmarks.
All of these experiments validate  \prunemethod{} and demonstrate the broad scope of enhanced performance across various settings.
In particular, we show that \prunemethod{} does not only improve in pruning performance, but it is also able to prune the models to a higher sparsity levels than other methods.
This robustness to high sparsities degrees is especially interesting since  model compression starts being useful when the reduction is at least one order of magnitude, i.e.,
at least 90\% sparsity. Lower sparsities (25\% or 37.5\%) even though interesting from a research perspective, are less applicable for  real world scenarios,
since the memory and speed-up benefits are not sufficient. 


This paper explores the short comings of current structured pruning methods in high sparsity regimes.
Even though, it is a new field, we hope that by exploring higher sparsity regimes
future research can develop pruning methods with gains comparable to Quantization and Knowledge Distillation.
This would open a new avenue to reduce the requirements of LLMs or possibly, improving current techniques by augmenting it with Structured Pruning.

\bibliographystyle{unsrt}
\bibliography{bibliography}





\appendix

\section{Appendix}

\subsection{Additional results}

As mentioned in the main paper, we delegate some of the results to this section due to space constrains.
We will in this section add an extensive list of all the experiment results.
To not extend the tables over the margin constrains we abbreviate sparsity with S.,
EvoPress with E.V., ShortGPT with Short. and BlockPruner with B.P.
Additionally, results over 10000 would be rounded to 10K notation.

\begin{table}
\centering
\begin{tabular}{llccccccccc}
\toprule
\multirow{2.5}*{Spars.} & \multirow{2.5}*{Method} & \multicolumn{3}{c}{Qwen3-0.6B} & \multicolumn{3}{c}{Qwen3-1.7B} & \multicolumn{3}{c}{Qwen3-4B} \\
\cmidrule(lr){3-5} \cmidrule(lr){6-8} \cmidrule(lr){9-11}
 &  & Wiki & FW & C4 & Wiki & FW & C4 & Wiki & FW & C4 \\
\midrule
0 & - & 20.97 & 20.69 & 29.45 & 16.75 & 16.11 & 22.28 & 13.75 & 13.54 & 20.05 \\
\midrule
\multirow{6}*{0.25} & UIDL & 23K & 21K & 33K & $\infty$ & $\infty$ & $\infty$ & 1994 & 555.5 & 771.5 \\
 & Short & 189.1 & 146.6 & 187.6 & 104.8 & 103.2 & 134.1 & 48.56 & 43.28 & 62.34 \\
 & B.P. & 41.28 & \underline{38.41} & \underline{52} & 30.86 & 28.94 & 39.03 & \underline{20.41} & 21.02 & 29.56 \\
 & E.P. & \underline{41.12} & 39.88 & 53.75 & 28.33 & 27.67 & 37.91 & 22.67 & 22.02 & 30.5 \\
 & 2SSP & 48 & 42.19 & 56.56 & \underline{26.41} & \underline{26.66} & \underline{34.38} & 20.48 & \underline{18.44} & \underline{24.75} \\
\rowcolor{gray!50} \cellcolor{white} & Putri & \textbf{35.03} & \textbf{32.66} & \textbf{42.59} & \textbf{22.11} & \textbf{23.77} & \textbf{28.83} & \textbf{15.77} & \textbf{17.38} & \textbf{22.36} \\
\midrule
\multirow{6}*{0.375} & UIDL & $\infty$ & $\infty$ & $\infty$ & 4544 & 2704 & 4688 & 828 & 733.5 & 831 \\
 & Short & 5568 & 2790 & 3710 & 54944 & 23440 & 47744 & 198.1 & 189.1 & 236.2 \\
 & B.P. & \underline{110.8} & 102 & 131 & 159.2 & 90.38 & 125.9 & 51.78 & 51 & 70.94 \\
 & E.P. & 125.4 & 103.2 & 136.2 & 106.1 & 74.31 & 101.6 & 44.31 & 42.03 & 58.34 \\
 & 2SSP & 112.4 & \underline{94.69} & \underline{118.3} & \underline{50.59} & \underline{46.69} & \underline{54.19} & \underline{35.19} & \underline{28.72} & \underline{35.75} \\
\rowcolor{gray!50} \cellcolor{white} & Putri & \textbf{52.19} & \textbf{51.28} & \textbf{62.59} & \textbf{32.78} & \textbf{32.28} & \textbf{36.22} & \textbf{23.86} & \textbf{26.14} & \textbf{30.62} \\
\midrule
\multirow{6}*{0.5} & UIDL & $\infty$ & $\infty$ & $\infty$ & $\infty$ & $\infty$ & $\infty$ & 3768 & 2946 & 3950 \\
 & Short & $\infty$ & $\infty$ & $\infty$ & 58944 & 11072 & 12848 & 2570 & 1499 & 1780 \\
 & B.P. & 409.8 & 328 & 398.8 & 923.5 & 446.5 & 566.5 & 143.2 & 120.7 & 163.6 \\
 & E.P. & \underline{356} & 272 & 358.8 & 7792 & 5976 & 6664 & 117.9 & 100 & 136.2 \\
 & 2SSP & 371.8 & \underline{255.4} & \underline{311.8} & \underline{166.2} & \underline{109.9} & \underline{112.4} & \underline{63.84} & \underline{56.56} & \underline{62.59} \\
\rowcolor{gray!50} \cellcolor{white} & Putri & \textbf{108.2} & \textbf{97.69} & \textbf{112} & \textbf{65.62} & \textbf{55.47} & \textbf{58.34} & \textbf{40.97} & \textbf{36.44} & \textbf{42.19} \\
\midrule
\multirow{6}*{0.75} & UIDL & $\infty$ & $\infty$ & $\infty$ & $\infty$ & $\infty$ & $\infty$ & $\infty$ & $\infty$ & $\infty$ \\
 & Short & $\infty$ & $\infty$ & $\infty$ & $\infty$ & $\infty$ & $\infty$ & $\infty$ & $\infty$ & $\infty$ \\
 & B.P. & 21K & 13K & 14K & $\infty$ & $\infty$ & $\infty$ & 15K & 10K & 15K \\
 & E.P. & 6716 & 4952 & 5792 & 9544 & 6984 & 7852 & 23K & 15K & 23K \\
 & 2SSP & \underline{2866} & \underline{2198} & \underline{2294} & \underline{1080} & \underline{768.5} & \underline{790} & \underline{1596} & \underline{1063} & \underline{1042} \\
\rowcolor{gray!50} \cellcolor{white} & Putri & \textbf{1441} & \textbf{1038} & \textbf{1114} & \textbf{884.5} & \textbf{647} & \textbf{649.5} & \textbf{309.2} & \textbf{207.6} & \textbf{224.5} \\
\midrule
\multirow{6}*{0.9} & UIDL & \underline{$\infty$} & $\infty$ & $\infty$ & $\infty$ & $\infty$ & $\infty$ & $\infty$ & $\infty$ & $\infty$ \\
 & Short & \underline{$\infty$} & $\infty$ & $\infty$ & $\infty$ & $\infty$ & $\infty$ & $\infty$ & $\infty$ & $\infty$ \\
 & B.P. & nan & nan & nan & nan & nan & nan & nan & nan & nan \\
 & E.P. & \underline{$\infty$} & \underline{65K} & \underline{64K} & 52K & \underline{24K} & \underline{30K} & \underline{58K} & \underline{55K} & \underline{61K} \\
 & 2SSP & nan & nan & nan & \underline{30816} & 58K & 42K & nan & nan & nan \\
\rowcolor{gray!50} \cellcolor{white} & Putri & \textbf{4440} & \textbf{2414} & \textbf{2404} & \textbf{2610} & \textbf{1685} & \textbf{1705} & \textbf{916.5} & \textbf{705.5} & \textbf{711} \\
\midrule
\multirow{6}*{0.95} & UIDL & \textbf{$\infty$} & \textbf{$\infty$} & \textbf{$\infty$} & \underline{$\infty$} & \underline{$\infty$} & \underline{$\infty$} & \underline{$\infty$} & \underline{$\infty$} & \underline{$\infty$} \\
 & Short. & \textbf{$\infty$} & \textbf{$\infty$} & \textbf{$\infty$} & \underline{$\infty$} & \underline{$\infty$} & \underline{$\infty$} & \underline{$\infty$} & \underline{$\infty$} & \underline{$\infty$} \\
 & B.P. & nan & nan & nan & nan & nan & nan & nan & nan & nan \\
 & E.P. & \textbf{$\infty$} & \textbf{$\infty$} & \textbf{$\infty$} & \underline{$\infty$} & \underline{$\infty$} & \underline{$\infty$} & \underline{$\infty$} & \underline{$\infty$} & \underline{$\infty$} \\
 & 2SSP & nan & nan & nan & nan & nan & nan & nan & nan & nan \\
\rowcolor{gray!50} \cellcolor{white} & Putri & \textbf{$\infty$} & \textbf{$\infty$} & \textbf{$\infty$} & \textbf{9400} & \textbf{6068} & \textbf{6116} & \textbf{3596} & \textbf{2258} & \textbf{2156} \\
\midrule
\multirow{6}*{0.99} & UIDL & \underline{$\infty$} & \underline{$\infty$} & \underline{$\infty$} & \textbf{$\infty$} & \textbf{$\infty$} & \textbf{$\infty$} & \textbf{$\infty$} & \textbf{$\infty$} & \textbf{$\infty$} \\
 & Short & \underline{$\infty$} & \underline{$\infty$} & \underline{$\infty$} & \textbf{$\infty$} & \textbf{$\infty$} & \textbf{$\infty$} & \textbf{$\infty$} & \textbf{$\infty$} & \textbf{$\infty$} \\
 & B.P. & nan & nan & nan & nan & nan & nan & nan & nan & nan \\
 & E.P. & nan & nan & nan & nan & nan & nan & nan & nan & nan \\
 & 2SSP & nan & nan & nan & nan & nan & nan & nan & nan & nan \\
\rowcolor{gray!50} \cellcolor{white} & Putri & \textbf{64K} & \textbf{29K} & \textbf{27K} & \textbf{$\infty$} & \textbf{$\infty$} & \textbf{$\infty$} & \textbf{$\infty$} & \textbf{$\infty$} & \textbf{$\infty$} \\
\bottomrule
\end{tabular}
\end{table}

\begin{table}
\centering
\begin{tabular}{llccccccccc}
\toprule
\multirow{2.5}*{Spars.} & \multirow{2.5}*{Meth.} & \multicolumn{3}{c}{Qwen3-8B} & \multicolumn{3}{c}{Qwen3-14B} & \multicolumn{3}{c}{Qwen3-32B} \\
\cmidrule(lr){3-5} \cmidrule(lr){6-8} \cmidrule(lr){9-11}
 &  & Wiki & FW & C4 & Wiki & FW & C4 & Wiki & FW & C4 \\
\midrule
0 & - & 9.734 & 10.48 & 15.19 & 8.641 & 9.43 & 13.73 & 7.594 & 8.32 & 12.26 \\
\midrule
\multirow{6}*{0.25} & UIDL & 5072 & 739 & 551.5 & 23.03 & 24.28 & 34.5 & $\infty$ & $\infty$ & $\infty$ \\
 & Short. & 38.56 & 33.91 & 46.16 & 39.56 & 38.12 & 52.19 & 14.16 & 15.31 & 23.2 \\
 & B.P. & 22.67 & 18.66 & 25.59 & 16.98 & 17.28 & 24.23 & 10.06 & 11.34 & 16.14 \\
 & E.P. & 22.02 & 18.66 & 26 & 14.95 & 15.19 & 21.3 & OOM & OOM & OOM \\
 & 2SSP & \underline{14.49} & \underline{16.17} & \underline{20.92} & \underline{11.51} & \underline{12.91} & \underline{16.88} & \underline{10.02} & \underline{11.31} & \underline{15.34} \\
 \rowcolor{gray!50} \cellcolor{white} & Putri & \textbf{13.54} & \textbf{15.83} & \textbf{19.2} & \textbf{10.62} & \textbf{12.28} & \textbf{15.46} & \textbf{9.13} & \textbf{13.91} & \textbf{10.48} \\
\midrule
\multirow{6}*{0.375} & UIDL & 36320 & $\infty$ & $\infty$ & 30336 & 15376 & 16368 & $\infty$ & $\infty$ & $\infty$ \\
 & Short. & 246.6 & 145 & 182.5 & 151.4 & 131 & 161.8 & 28.44 & 29.8 & 42.09 \\
 & B.P. & 47.53 & 39.31 & 53.03 & 46.78 & 39.47 & 54.81 & 14.22 & 15.58 & 22.06 \\
 & E.P. & 42.59 & 38.25 & 51.19 & 25.34 & 24.19 & 34.03 & OOM & OOM & OOM \\
 & 2SSP & \underline{21.12} & \underline{24.42} & \underline{28.66} & \underline{15.25} & \underline{16.98} & \underline{21.05} & \underline{13.43} & \underline{15.31} & \underline{19.47} \\
 \rowcolor{gray!50} \cellcolor{white} & Putri & \textbf{18.25} & \textbf{20.2} & \textbf{23.3} & \textbf{13.51} & \textbf{15.61} & \textbf{18.28} & \textbf{11.03} & \textbf{15.46} & \textbf{12.28} \\
\midrule
\multirow{6}*{0.5} & UIDL & $\infty$ & $\infty$ & $\infty$ & $\infty$ & $\infty$ & $\infty$ & $\infty$ & $\infty$ & $\infty$ \\
 & Short. & 1787 & 639.5 & 655 & 1318 & 622.5 & 697 & $\infty$ & $\infty$ & $\infty$ \\
 & B.P. & 202.9 & 108.2 & 143.2 & 221 & 131.5 & 181.9 & 39.25 & 35.03 & 48 \\
 & E.P. & 211 & 120.7 & 151.9 & 86.94 & 73.75 & 101.6 & OOM & OOM & OOM \\
 & 2SSP & \underline{41.94} & \underline{40.09} & \underline{44.22} & \underline{23.53} & \underline{25} & \underline{29.75} & \underline{19.39} & \underline{21.67} & \underline{26.56} \\
 \rowcolor{gray!50} \cellcolor{white} & Putri & \textbf{28.44} & \textbf{27.3} & \textbf{30.09} & \textbf{18.19} & \textbf{19.97} & \textbf{22.98} & \textbf{15.16} & \textbf{16.95} & \textbf{20.2} \\
\midrule
\multirow{6}*{0.75} & UIDL & $\infty$ & $\infty$ & $\infty$ & $\infty$ & $\infty$ & $\infty$ & $\infty$ & $\infty$ & $\infty$ \\
 & Short. & $\infty$ & $\infty$ & $\infty$ & $\infty$ & $\infty$ & $\infty$ & 65K & $\infty$ & $\infty$ \\
 & B.P. & 13K & 5884 & 5884 & 5524 & 4336 & 4544 & 5884 & 3224 & 3568 \\
 & E.P. & 5112 & 3768 & 4042 & 7208 & 4840 & 5316 & OOM & OOM & OOM \\
 & 2SSP & \underline{536.5} & \underline{395.8} & \underline{385} & \underline{212.6} & \underline{160.5} & \underline{151.9} & \underline{122.1} & \underline{99.62} & \underline{94.31} \\
 \rowcolor{gray!50} \cellcolor{white} & Putri & \textbf{164.2} & \textbf{140} & \textbf{149.6} & \textbf{124.5} & \textbf{68.75} & \textbf{72.31} & \textbf{44.66} & \textbf{41.78} & \textbf{44.56} \\
\midrule
\multirow{6}*{0.9} & UIDL & $\infty$ & $\infty$ & $\infty$ & $\infty$ & $\infty$ & $\infty$ & nan & nan & nan \\
 & Short. & $\infty$ & $\infty$ & $\infty$ & $\infty$ & $\infty$ & $\infty$ & nan & nan & nan \\
 & B.P. & nan & nan & nan & nan & nan & nan & 27K & 28K & 28K \\
 & E.P. & 34K & 27K & 27K & 9928 & 7208 & 7976 & OOM & OOM & OOM \\
 & 2SSP & \underline{12K} & \underline{9256} & \underline{7436} & \underline{5316} & \underline{2662} & \underline{2432} & \underline{2866} & \underline{1476} & \underline{1282} \\
 \rowcolor{gray!50} \cellcolor{white} & Putri & \textbf{1030} & \textbf{617.5} & \textbf{634.5} & \textbf{1307} & \textbf{716.5} & \textbf{673} & \textbf{370.2} & \textbf{215.1} & \textbf{214.2} \\
\midrule
\multirow{6}*{0.95} & UIDL & \underline{$\infty$} & \underline{$\infty$} & \underline{$\infty$} & $\infty$ & $\infty$ & $\infty$ & $\infty$ & $\infty$ & $\infty$ \\
 & Short. & \underline{$\infty$} & \underline{$\infty$} & \underline{$\infty$} & $\infty$ & $\infty$ & $\infty$ & $\infty$ & $\infty$ & $\infty$ \\
 & B.P. & nan & nan & nan & nan & nan & nan & nan & nan & nan \\
 & E.P. & \underline{$\infty$} & \underline{$\infty$} & \underline{$\infty$} & \underline{11K} & \underline{8360} & \underline{9328} & OOM & OOM & OOM \\
 & 2SSP & \underline{$\infty$} & \underline{$\infty$} & \underline{$\infty$} & nan & nan & nan & \underline{30K} & \underline{23K} & \underline{25K} \\
 \rowcolor{gray!50} \cellcolor{white} & Putri & \textbf{3124} & \textbf{1540} & \textbf{1571} & \textbf{4916} & \textbf{1880} & \textbf{1583} &  \textbf{2714} & \textbf{831} & \textbf{871} \\
\midrule
\multirow{6}*{0.99} & UIDL & \textbf{$\infty$} & \textbf{$\infty$} & \textbf{$\infty$} & nan & nan & nan & nan & nan & nan \\
 & Short. & \textbf{$\infty$} & \textbf{$\infty$} & \textbf{$\infty$} & $\infty$ & $\infty$ & $\infty$ & $\infty$ & $\infty$ & $\infty$ \\
 & B.P. & nan & nan & nan & nan & nan & nan & nan & nan & nan \\
 & E.P. & nan & nan & nan & nan & nan & nan & OOM & OOM & OOM \\
 & 2SSP & nan & nan & nan & nan & nan & nan & nan & nan & nan \\
 \rowcolor{gray!50} \cellcolor{white} & Putri & nan & nan & nan & \textbf{47K} & \textbf{40K} & \textbf{39K} &  $\infty$ & $\infty$ & $\infty$ \\
\bottomrule
\end{tabular}
\end{table}

\begin{table}
\centering
\begin{tabular}{llcccccc}
\toprule
\multirow{2.5}*{Sparsity} & \multirow{2.5}*{Method} & \multicolumn{3}{c}{Meta-Llama-3-8B} & \multicolumn{3}{c}{Llama-3.1-8B} \\
\cmidrule(lr){3-5} \cmidrule(lr){6-8}
 &  & Wiki & FW & C4 & Wiki & FW & C4 \\
\midrule
0 & - & 6.312 & 8.5 & 8.125 & 6.406 & 8.375 & 8.25 \\
\midrule
\multirow{6}*{0.25} & UIDL & 148 & 96 & 153 & 2976 & 3168 & 2976 \\
 & Short. & 223 & 174 & 191 & 260 & 296 & 278 \\
 & B.P. & 14.44 & 17.5 & 20.38 & 14.69 & 18 & 21 \\
 & E.P. & 12.94 & \underline{14.44} & \underline{18.88} & 13.56 & 15.62 & 20.12 \\
 & 2SSP & \textbf{11.81} & \textbf{14} & \textbf{16.62} & \underline{12} & \underline{14.44} & \underline{16.62} \\
 \rowcolor{gray!50} \cellcolor{white} & Putri & \underline{12.38} & 19.75 & 21.75 & \textbf{11.81} & \textbf{13.38} & \textbf{14.94} \\
\midrule
\multirow{6}*{0.375} & UIDL & 392 & 286 & 368 & 111616 & 20736 & 15168 \\
 & Short. & 1408 & 856 & 968 & 8640 & 18304 & 14208 \\
 & B.P. & 32 & 32 & 43.25 & 27 & 29.25 & 37.5 \\
 & E.P. & 31.12 & 29.25 & 39.25 & 26.25 & 27 & 34.25 \\
 & 2SSP & \underline{21.38} & \underline{22} & \underline{24.25} & \underline{20.38} & \underline{21.75} & \underline{23.5} \\
 \rowcolor{gray!50} \cellcolor{white} & Putri & \textbf{16.88} & \textbf{19.5} & \textbf{20.75} & \textbf{17.5} & \textbf{19.75} & \textbf{21.75} \\
\midrule
\multirow{6}*{0.5} & UIDL & 1000 & 688 & 732 & 52736 & 38656 & 34048 \\
 & Short. & 56320 & 49664 & 46592 & 14208 & 10432 & 8640 \\
 & B.P. & 131 & 99 & 108.5 & 163 & 163 & 185 \\
 & E.P. & 96 & 99 & 108.5 & 148 & 96 & 105 \\
 & 2SSP & \underline{34.75} & \textbf{40.5} & \underline{46.75} & \textbf{34.75} & \textbf{41.25} & \underline{47.5} \\
 \rowcolor{gray!50} \cellcolor{white} & Putri & \textbf{32} & \underline{45.25} & \textbf{45.25} & \underline{39.25} & \underline{41.75} & \textbf{46} \\
\midrule
\multirow{6}*{0.75} & UIDL & 17152 & 15168 & 17152 & 344064 & 196608 & 391168 \\
 & Short. & 1204224 & 1359872 & 2113536 & 49664 & 49664 & 56320 \\
 & B.P. & 7136 & 8640 & 6720 & 98816 & 28288 & 26624 \\
 & E.P. & 20736 & 8640 & 7616 & 17152 & 8640 & 8640 \\
 & 2SSP & \underline{346} & \underline{268} & \underline{268} & \underline{278} & \underline{260} & \underline{252} \\
 \rowcolor{gray!50} \cellcolor{white} & Putri & \textbf{216} & \textbf{237} & \textbf{223} & \textbf{163} & \textbf{185} & \textbf{203} \\
\midrule
\multirow{6}*{0.9} & UIDL & 880640 & 8912896 & 8912896 & 1859584 & 21364736 & 10092544 \\
 & Short. & 323584 & 152576 & 162816 & 1859584 & 21364736 & 10092544 \\
 & B.P. & 119296 & 184320 & 126976 & 501760 & 684032 & 643072 \\
 & E.P. & 46592 & 41216 & 38656 & 28288 & 14208 & 20736 \\
 & 2SSP & \underline{1592} & \underline{1032} & \underline{940} & \underline{1544} & \underline{908} & \underline{908} \\
 \rowcolor{gray!50} \cellcolor{white} & Putri & \textbf{568} & \textbf{458} & \textbf{472} & \textbf{644} & \textbf{416} & \textbf{472} \\
\midrule
\multirow{6}*{0.95} & UIDL & 569344 & 501760 & 442368 & 501760 & 501760 & 415744 \\
 & Short. & 251904 & 143360 & 184320 & 532480 & 391168 & 471040 \\
 & B.P. & 827392 & 778240 & 827392 & 1204224 & 995328 & 995328 \\
 & E.P. & 92672 & 68096 & 72192 & 135168 & 98816 & 98816 \\
 & 2SSP & \underline{4928} & \underline{2976} & \underline{2976} & \underline{8096} & \underline{2976} & \underline{2976} \\
 \rowcolor{gray!50} \cellcolor{white} & Putri & \textbf{2048} & \textbf{828} & \textbf{880} & \textbf{1696} & \textbf{1096} & \textbf{968} \\
\midrule
\multirow{6}*{0.99} & UIDL & $\infty$ & $\infty$ & $\infty$ & -- & -- & -- \\
 & Short. & 569344 & 569344 & 569344 & 532480 & 501760 & 442368 \\
 & B.P. & 569344 & 569344 & 569344 & 532480 & 501760 & 442368 \\
 & E.P. & nan & nan & nan & nan & nan & nan \\
 & 2SSP & \textbf{22016} & \underline{14208} & \underline{17152} & \underline{34048} & \underline{17152} & \underline{17152} \\
 \rowcolor{gray!50} \cellcolor{white} & Putri & \underline{24960} & \textbf{9152} & \textbf{8096} & \textbf{11776} & \textbf{5216} & \textbf{3824} \\
\bottomrule
\end{tabular}
\end{table}

\begin{table}
\centering
\begin{tabular}{llcccccc}
\toprule
\multirow{2.5}*{Sparsity} & \multirow{2.5}*{Method} & \multicolumn{3}{c}{Llama-3.2-1B} & \multicolumn{3}{c}{Llama-3.2-3B} \\
\cmidrule(lr){3-5} \cmidrule(lr){6-8}
 &  & Wiki & FW & C4 & Wiki & FW & C4 \\
\midrule
0 & - & 10.25 & 12 & 12.94 & 8.375 & 10.12 & 10.75 \\
\midrule
\multirow{6}*{0.25} & UIDL & 7616 & 2976 & 4608 & 1496 & 908 & 1280 \\
 & Short. & 245 & 237 & 245 & 158 & 168 & 203 \\
 & B.P. & 42.5 & 38.75 & 54.5 & 23.5 & 22.75 & 29.25 \\
 & E.P. & 51.25 & 48.25 & 54.5 & 22.75 & \underline{21.38} & \underline{26.62} \\
 & 2SSP & \underline{18.62} & \underline{20.12} & \underline{25.38} & \textbf{16.12} & \textbf{19.5} & \textbf{20.75} \\
 \rowcolor{gray!50} \cellcolor{white} & Putri & \textbf{16.38} & \textbf{19.5} & \textbf{21.75} & \underline{16.88} & 23.12 & 27.5 \\
\midrule
\multirow{6}*{0.375} & UIDL & 52736 & 34048 & 34048 & 7616 & 5568 & 5568 \\
 & Short. & 260 & 191 & 209 & 458 & 416 & 486 \\
 & B.P. & 203 & 174 & 191 & 53.75 & 51.25 & \underline{54.5} \\
 & E.P. & 163 & 119.5 & 148 & 50.5 & 46.75 & \underline{54.5} \\
 & 2SSP & \underline{31.62} & \underline{35.75} & \underline{50.5} & \textbf{26.62} & \textbf{29.25} & \textbf{32.5} \\
 \rowcolor{gray!50} \cellcolor{white} & Putri & \textbf{24.62} & \textbf{26.62} & \textbf{30.12} & \underline{31.62} & \underline{45.25} & \underline{54.5} \\
\midrule
\multirow{6}*{0.5} & UIDL & 222208 & 162816 & 173056 & 46592 & 34048 & 36352 \\
 & Short. & 3168 & 1592 & 1592 & 11072 & 9792 & 11776 \\
 & B.P. & 1592 & 968 & 1032 & 139 & 127 & 153 \\
 & E.P. & 536 & 392 & 404 & 163 & 135 & 153 \\
 & 2SSP & \underline{56.25} & \underline{79.5} & \underline{77} & \underline{49.75} & \underline{49} & \underline{54.5} \\
 \rowcolor{gray!50} \cellcolor{white} & Putri & \textbf{35.75} & \textbf{38.75} & \textbf{53.75} & \textbf{40} & \textbf{40.5} & \textbf{48.25} \\
\midrule
\multirow{6}*{0.75} & UIDL & 778240 & 684032 & 880640 & 152576 & 126976 & 126976 \\
 & Short. & 442368 & 1359872 & 6094848 & 38656 & 14208 & 19456 \\
 & B.P. & 14208 & 8096 & 7136 & 4928 & 2976 & 2976 \\
 & E.P. & 28288 & 18304 & 28288 & 7616 & 5216 & 5568 \\
 & 2SSP & \underline{472} & \underline{368} & \underline{404} & \underline{520} & \underline{378} & \underline{378} \\
 \rowcolor{gray!50} \cellcolor{white} & Putri & \textbf{158} & \textbf{209} & \textbf{216} & \textbf{191} & \textbf{209} & \textbf{216} \\
\midrule
\multirow{6}*{0.9} & UIDL & $\infty$ & $\infty$ & $\infty$ & $\infty$ & $\infty$ & $\infty$ \\
 & Short. & $\infty$ & $\infty$ & $\infty$ & $\infty$ & $\infty$ & $\infty$ \\
 & B.P. & $\infty$ & $\infty$ & $\infty$ & 152576 & 173056 & 173056 \\
 & E.P. & 49664 & 56320 & 59904 & 12544 & 8640 & 8096 \\
 & 2SSP & \underline{2976} & \underline{1408} & \underline{1456} & \underline{2464} & \underline{1368} & \underline{1320} \\
 \rowcolor{gray!50} \cellcolor{white} & Putri & \textbf{1096} & \textbf{856} & \textbf{752} & \textbf{752} & \textbf{604} & \textbf{664} \\
\midrule
\multirow{6}*{0.95} & UIDL & $\infty$ & $\infty$ & $\infty$ & $\infty$ & $\infty$ & $\infty$ \\
 & Short. & $\infty$ & $\infty$ & $\infty$ & $\infty$ & $\infty$ & $\infty$ \\
 & B.P. & nan & nan & nan & $\infty$ & $\infty$ & $\infty$ \\
 & E.P. & nan & nan & nan & 501760 & 143360 & 126976 \\
 & 2SSP & \underline{17152} & \underline{19456} & \underline{14208} & \underline{5568} & \underline{2896} & \underline{2976} \\
 \rowcolor{gray!50} \cellcolor{white} & Putri & \textbf{2624} & \textbf{1320} & \textbf{1240} & \textbf{1984} & \textbf{1240} & \textbf{1000} \\
\midrule
\multirow{6}*{0.99} & UIDL & $\infty$ & \textbf{$\infty$} & \textbf{$\infty$} & $\infty$ & $\infty$ & $\infty$ \\
 & Short. & $\infty$ & \textbf{$\infty$} & \textbf{$\infty$} & $\infty$ & $\infty$ & $\infty$ \\
 & B.P. & nan & nan & nan & $\infty$ & $\infty$ & $\infty$ \\
 & E.P. & nan & nan & nan & nan & nan & nan \\
 & 2SSP & nan & nan & nan & \underline{13376} & \underline{11776} & \underline{11776} \\
 \rowcolor{gray!50} \cellcolor{white} & Putri & nan & nan & nan & \textbf{10432} & \textbf{5920} & \textbf{5920} \\
\bottomrule
\end{tabular}
\end{table}


\end{document}